# Sparse matrix-variate Gaussian process blockmodels for network modeling

**Feng Yan**
Computer Science Dept.
Purdue University
West Lafayette, IN 47907, USA

**Zenglin Xu**
Computer Science Dept.
Purdue University
West Lafayette, IN 47907, USA

**Yuan Qi**
Computer Science and Statistics Depts.
Purdue University
West Lafayette, IN 47907, USA

## Abstract

We face network data from various sources, such as protein interactions and online social networks. A critical problem is to model network interactions and identify latent groups of network nodes. This problem is challenging due to many reasons. For example, the network nodes are interdependent instead of independent of each other, and the data are known to be very noisy (e.g., missing edges). To address these challenges, we propose a new relational model for network data, Sparse Matrix-variate Gaussian process Blockmodel (SMGB). Our model generalizes popular bilinear generative models and captures nonlinear network interactions using a matrix-variate Gaussian process with latent membership variables. We also assign sparse prior distributions on the latent membership variables to learn sparse group assignments for individual network nodes. To estimate the latent variables efficiently from data, we develop an efficient variational expectation maximization method. We compared our approaches with several state-of-the-art network models on both synthetic and real-world network datasets. Experimental results demonstrate SMGBs outperform the alternative approaches in terms of discovering latent classes or predicting unknown interactions.

## 1 Introduction

Networks are ubiquitous. Proteins interact with each other and form protein-protein interaction networks; and researchers write papers collaboratively and create co-author networks. Given network datasets with ever-increasing sizes from various sources, a common task is to model network interactions and to identify latent groups among network nodes. For instance, we may want to discover groups of researchers who share common research interests in a co-author network, or to identify modules in a protein-interaction network and predict possible missing interactions.

This task presents new modeling challenges however. First, we cannot use classical independence assumptions for network data analysis. The objects are interdependent via interactions or links between them, necessitating new models that capture relations among objects. Second, relationships among objects and latent groups may be quite complicated. A simple linear—or bilinear—model may not be sufficient to model the complex relationships. Third, network data is noisy. For example, protein-interaction data are known to have a limited coverage of all interactions and contain many false positives in detected interactions.

To address these challenges, we propose a Sparse Matrix-variate Gaussian process Blockmodel (SMGB). To model nonlinear interactions between network nodes and latent groups, the SMGB uses nonlinear Gaussian covariance functions, leading to equivalent nonlinear matrix factorization. SMGB naturally generalizes Gaussian-process latent variable models (GP-LVMs) (Lawrence and Urtasun, 2009) in multiple ways. First, Gaussian process latent variable models assume objects, i.e., corresponding to columns of a data matrix, are independent of each other. By contrast, our model uses both row- and column-wise covariance functions to represent row- and column-wise interactions, capturing interdependence between objects. Second, while GP-LVMs focus on Gaussian regression likelihoods, our models handles binary (and other more complex) interactions specially designed for discrete network data. Third, unlike GP-LVMs, our model uses a sparsity prior to encourage sparse memberships of individual network nodes.

To efficiently learn SMGB , we propose a novel varia-tional inference technique. Classical Gaussian process inference techniques requires $O(n^6)$ time complexity and $O(n^4)$ space complexity. Our algorithm reduce the time complexity and the space complexity of exact inference to $O(n^3)$ and $O(n^2)$, respectively. The complexities can be further reduced by approximating the covariance function by truncated singular value decomposition (SVD).

In section 2 we will describe related work in detail. Section 3 and 4 present the novel SMGBs and the efficient variational expectation maximization algorithm for efficient model estimation. The experimental results in Section 5 show that the new SMGBs outperform the state-of-the-art network models, such as mixed membership stochastic blockmodels (Airoldi et al., 2008) and latent eigenmodels (Hoff, 2007) in terms of discovering latent classes or predicting unknown interactions.

## 2 Related Work

Due to the importance of network modeling, various approaches have been proposed. Many of them aim to model stochastic equivalence between network nodes and accordingly divide the nodes into latent classes. The latent classes provide building blocks for complex networks and allow us to understand and predict unknown interactions between network nodes. For example, stochastic blockmodels (Snijders and Nowicki, 1997) adopt mixture models for relational data. In this model, each node is sampled from a cluster based on a multinomial distribution. To avoid a predefined number of clusters, Kemp et al. (2004) replaced the multinomial distribution on latent membership variables by a flexible Chinese restaurant process prior. To allow a node belonging to multiple groups, Airoldi et al. (2008) developed *mixed* membership stochastic blockmodels, which use a latent Dirichlet allocation prior to model latent membership variables. Recently Hoff (2007) provided a unified view to latent class models and latent distance models (Hoff et al., 2002) based on eigenmodels. All these approaches differ in their priors on membership variables, but they all essentially represent the relationships between nodes by a bilinear model that connects membership variables via a binary or continuous or diagonal interaction matrix.

The bilinear model, however, may not be sufficient to capture complex social and biological interactions. Our model essentially generalizes the bilinear generative models to handle nonlinear network interactions. Grounded in a nonparametric Bayesian framework, this generalization is achieved by the adoption of relational Gaussian processes. Furthermore, unlike the latent eigenmodel, the SMGB uses a *sparse* prior distribution on the membership variables, so that we can easily interpret sparse group assignments for individual network nodes. Our approach also differs from the eigenmodel approach in terms of estimation methods: while the latent eigenmodel is estimated by Monte Carlo methods (Hoff, 2007), we use variational EM to estimate the latent memberships and predict unknown interactions (See Section 4 for details).

Our model is also related to GP-LVMs (Lawrence, 2006, Lawrence and Urtasun, 2009). The critical differences have been discussed in the previous section, e.g., the relational modeling of objects by our model vs. the independent modeling of objects by GP-LVMs. In addition, our model is closely related to the relational Gaussian process model (Yu and Chu, 2008), which has been successfully applied to a variety of relational learning problems. Essentially Yu et al. use a linear covariance function and Gaussian likelihood in their relational GP model. By contrast, we use a nonlinear covariance function for relational GP, coupled with sparse priors on latent membership variables, and probit likelihoods. These differences not only make SMGB suitable for discrete network data, but also allow us to learn low-dimensional sparse memberships, while the previous approach cannot. Recently, Xu et al. (2011) propose a similar model based on matrix-variate $t$-process, which only use the Gaussian noise model.

## 3 Sparse matrix-variate Gaussian process blockmodels (SMGBs)

In this section we present SMGB . First, we use an $n$ by $n$ interaction matrix $\mathbf{Y}$ to represent the *noisy* binary relationships between $n$ network entities. If $y_{ij} = 1$ where $y_{ij}$ is the $\{i, j\}$ element of $\mathbf{Y}$, there is an observed interaction between entities $i$ and $j$; and if $y_{ij} = 0$, there is no interaction between them. Given the observed entries in $\mathbf{Y}$, we want to use the SMGB to predict unknown entries in the network and estimate latent membership vector for each entity $i$. We denote this $d$ by 1 vector as $\mathbf{u}_i$, where $d$ is a given number of latent classes, and define the membership matrix $\mathbf{U} = (\mathbf{u}_1, \ldots, \mathbf{u}_n) \in \mathbb{R}^{d\times n}$.

### 3.1 Probit noise function

We assume the elements of the matrix $\mathbf{Y}$ are conditionally independent given a latent matrix $\mathbf{X} = \{x_{ij}\}$. These two matrices are linked together via probit functions:

$$p(\mathbf{Y}|\mathbf{X}) = \prod_{1\le i,j\le n} \Phi(x_{ij})^{y_{ij}}(1-\Phi(x_{ij}))^{(1-y_{ij})} \qquad (1)$$

where $\Phi(\cdot)$ is the cumulated distribution function of a standard normal distribution.

Although we focus on binary interactions in this paper, probit functions are capable of modeling multiple relationships between two network entities, either ordinal or unordered (Albert and Chib, 1993).

In this paper, we use an equivalent augmented representation of the probit model. Specifically, we follow Albert and Chib (1993) to introduce a latent matrix $\mathbf{Z} = \{z_{ij}\}$:

$$p(y_{ij}|z_{ij}, x_{ij}) = \{\delta(y_{ij} = 1)\delta(z_{ij} > 0) + \delta(y_{ij} = 0)\delta(z_{ij} \leq 0)\} \quad (2)$$

$$p(z_{ij} \mid x_{ij}) = \mathcal{N}(z_{ij}; x_{ij}, 1) \quad (3)$$

where $\delta(\cdot)$ is the indicator function (its value is 1 if the statement inside is true, and 0 otherwise), and $\mathcal{N}(x; \mu, \sigma^2)$ is the univariate normal probability density function with mean $\mu$ and variance $\sigma^2$. It is easy to verify that if we marginalize out $z_{ij}$, we recover the original probit function $p(y_{ij}|x_{ij})$.

If we have the side information about the network interactions, we can utilize them in our model. Specifically, let us denote the side information about the interaction between entities $i$ and $j$ as a vector $\mathbf{r}_{ij}$. Then we can represent the latent variable $x_{ij}$ as a combination of a linear predictor $p_{ij} = \beta^\top \mathbf{r}_{ij}$ and the $\{i, j\}$ entry $m_{ij}$ of a latent interaction matrix $\mathbf{M}$.

$$x_{ij} = \beta^\top \mathbf{r}_{ij} + m_{ij} \quad (4)$$

We assign a normal prior distribution over the regression weights $\beta$. If the side information is unavailable, we simply have $\mathbf{X} = \mathbf{M}$.

The matrix $\mathbf{M}$ is used to capture the underlying interactions between network entities. In particular, it is sampled from a matrix-variate Gaussian process as described in the following section.

### 3.2 Matrix-variate Gaussian Processes

In this section, we show how to obtain matrix-variate Gaussian Processes via nonlinear matrix factorization models. First let us assume the latent interaction matrix

$$\mathbf{M} = \mathbf{U}^\top \mathbf{W} \mathbf{U} \quad (5)$$

Clearly, the $(i, j)$ latent interaction $m_{ij} = \mathbf{u}_i^\top \mathbf{W} \mathbf{u}_j$ such that it is determined by the class memberships of entities $i$ and $j$ and the matrix $\mathbf{W}$. The $\{h, g\}$ element of $\mathbf{W}$ can be interpreted as the interaction strength between latent classes $h$ and $g$.

Note that if we force $\mathbf{W}$ to be a diagonal matrix, then this model reduces to the eigenmodel (Hoff, 2007).

Now we assign a matrix-variate normal distribution on $\mathbf{W}$ (Gupta and Nagar, 2000):

$$\mathbf{W} \sim \mathcal{N}_{d,d}(\mathbf{W}; \mathbf{0}, \Sigma, \Omega). \quad (6)$$

where $\mathbf{0}$ is the zero mean matrix, and $\Sigma$ and $\Omega$ are row-wise and column-wise covariance matrices respectively.

To model nonlinear row-wise interactions, we replace the membership variables $\mathbf{u}_i$ by a nonlinear feature mapping $\phi(\mathbf{u}_i)$. If we set $\Sigma = \Omega = \mathbf{I}$, we obtain an equivalent row-wise covariance matrix $\mathbf{K}$ with a nonlinear covariance function $k(\mathbf{u}_i, \mathbf{u}_j) = \phi(\mathbf{u}_i)^\top \phi(\mathbf{u}_j)$. Similarly, we can define a column-wise covariance matrix $\mathbf{G}$ with a possibly nonlinear covariance function $g$. As a result, $\mathbf{M}$ follows a *matrix-variate* Gaussian process (Yu et al., 2007) defined as follows.

**Definition 1 (Matrix-variate Gaussian processes)** *A matrix-variate Gaussian process is a stochastic process whose projection on any finite locations* $\mathbf{U} = [\mathbf{u}_1^\top, \ldots, \mathbf{u}_n^\top]^\top$ *follows a matrix-variate normal distribution. Specifically, given* $\mathbf{U}$*, the zero mean matrix-variate Gaussian process on* $\mathbf{M}$ *has the form:*

$$p(\mathbf{M}|\mathbf{U}) = \mathcal{GP}_{n,n}(\mathbf{M}; 0, \mathbf{K}, \mathbf{G}) \quad (7)$$

$$= (2\pi)^{-n^2/2} \det(\mathbf{K})^{-n/2} \det(\mathbf{G})^{-n/2} \exp\{-\frac{1}{2} \operatorname{tr}(\mathbf{K}^{-1}\mathbf{M}\mathbf{G}^{-1}\mathbf{M}^\top)\} \quad (8)$$

*where* $k_{ij} = k(\mathbf{u}_i, \mathbf{u}_j)$ *and* $g_{ij} = g(\mathbf{u}_i, \mathbf{u}_j)$.

For simplicity we set $\mathbf{K} = \mathbf{G}$ for the rest of the paper. Note that the matrix-variate Gaussian processes satisfy the consistency condition for any valid stochastic processes; i.e., given the finite projection of the process, its marginal distribution in a subspace is the same as its direct projection onto the subspace.

### 3.3 Sparse prior on latent membership vectors

For easy model interpretation, we assign a Laplace prior $p(\mathbf{U}) = \prod_i \exp(-\lambda \|\mathbf{u}_i\|_1)$, encouraging sparsity in the membership variables $\mathbf{u}_i$.

### 3.4 Joint distribution of SMGB

Combining all the model components together, we have the joint distribution of the SMGB

$$p(\mathbf{Y}, \mathbf{Z}, \mathbf{M}, \beta, \mathbf{U}) = \prod_{1 \leq i,j \leq n} p(y_{ij}|z_{ij}, x_{ij}) p(z_{ij}|x_{ij}) \cdot p(\beta) p(\mathbf{M}|\mathbf{U}) p(\mathbf{U}) \quad (9)$$

Given the observed $\mathbf{Y}$, the inference task is to estimate all the latent variables in the model.

## 4 Inference

To estimate the membership matrix $\mathbf{U}$, we want to maximize $p(\mathbf{U}|\mathbf{Y}) \propto p(\mathbf{Y}|\mathbf{U})p(\mathbf{U})$ where $p(\mathbf{Y}|\mathbf{U})$ is the marginal model likelihood. To this end, an expectation maximization algorithm can be used. The E step computes the expected log probability of the joint model (9) over the posterior distribution $p(\mathbf{Z}, \mathbf{M}, \beta|\mathbf{Y}, \mathbf{U})$ and the M step optimizes the expected log probability of the joint model over $\mathbf{U}$. It is, however, intractable to compute the exact posterior distribution $p(\mathbf{Z}, \mathbf{M}, \beta|\mathbf{Y}, \mathbf{U})$ needed by the E-step. We can use Gibbs sampling method to replace the exact expectation computation. But the resulting Gibbs-EM algorithm is slow due to its sampling nature and and it is practically difficult to assess the convergence of the sampler. Therefore, we uses a variational approach to approximate the exact posterior posterior distribution and accordingly the overall algorithm becomes a variational EM.

### 4.1 Variational expectation maximization

Our variational-EM algorithm consists of a variational E-step and a gradient-based M-step. In the E-step, we approximate the posterior distribution $p(\mathbf{Z}, \mathbf{M}, \beta|\mathbf{Y}, \mathbf{U})$ by a fully factorized distribution

$$q(\mathbf{Z}, \mathbf{M}, \beta) = q(\mathbf{Z})q(\mathbf{M})q(\beta) \tag{10}$$

Variational inference minimizes the KL divergence between the approximate and the exact posteriors

$$\min_q \mathrm{KL}\left(q(\mathbf{Z})q(\mathbf{M})q(\beta) \| p(\mathbf{Z}, \mathbf{M}, \beta|\mathbf{Y}, \mathbf{U})\right) \tag{11}$$

For this minimization, we iteratively update $q(\mathbf{Z})$, $q(\mathbf{M})$, and $q(\beta)$. More specifically, using a coordinate descent algorithm, the variational approach updates one approximate distribution, e.g, $q(\mathbf{Z})$, in (10) at a time while having all the others fixed. The derivation of the specific updates is similar to that of (Girolami and Rogers, 2005).

Given the current $q(\mathbf{Z})$ and $q(\beta)$, we update $q(\mathbf{M})$

$$q(\mathrm{vec}(\mathbf{M}^\top)) = \mathcal{N}(\mathrm{vec}(\langle\mathbf{M}\rangle^\top), \mathbf{\Sigma}_\mathbf{M}) \tag{12}$$

$$\mathbf{\Sigma}_\mathbf{M} = \mathbf{K} \otimes \mathbf{K}(\mathbf{I} + \mathbf{K} \otimes \mathbf{K})^{-1} \tag{13}$$

$$\mathrm{vec}(\langle\mathbf{M}\rangle^\top) = \mathbf{\Sigma}_\mathbf{M}\ \mathrm{vec}(\langle\mathbf{Z}\rangle^\top - \langle\mathbf{P}\rangle^\top) \tag{14}$$

where $\mathbf{P} = \{\beta^\top \mathbf{r}_{ij}\}$ and $\langle\cdot\rangle$ denotes the expectation under the approximate posteriors.

The variational distribution $q(z_{ij})$ is a truncated normal distribution. We update it as follows,

$$q(z_{ij}) \propto \mathcal{N}(\langle m_{ij}\rangle + \langle\beta\rangle^\top \mathbf{r}_{ij}, 1)\ \delta(z_{ij} > 0). \tag{15}$$

Since the left side of the normal density function is truncated, we need to adjust the mean of the normal distribution

$$\langle x_{ij}\rangle = \langle m_{ij}\rangle + \langle\beta\rangle^\top \mathbf{r}_{ij} \tag{16}$$

$$\langle z_{ij}\rangle = \langle x_{ij}\rangle + \frac{(2y_{ij}-1)\mathcal{N}(\langle m_{ij}\rangle; 0, 1)}{\Phi((2y_{ij}-1)\langle m_{ij}\rangle)} \tag{17}$$

Finally, we update the normal variational distribution $\beta$ as follows,

$$q(\beta) = \mathcal{N}(\langle\beta\rangle, \mathbf{\Sigma}_\beta) \tag{18}$$

$$\mathbf{\Sigma}_\beta = \left(\sum_{i,j} \mathbf{r}_{ij}\mathbf{r}_{ij}{}^\top + \sigma_\beta^{-2}\mathbf{I}\right)^{-1} \tag{19}$$

$$\langle\beta\rangle = \mathbf{\Sigma}_\beta \cdot \sum_{i,j}(\langle z_{ij}\rangle - \langle m_{ij}\rangle)\mathbf{r}_{ij} \tag{20}$$

We loop over the updates above until convergence to obtain the final variational approximate distributions.

Based on the variational distributions, we maximize the expected log-probability of the joint model over $\mathbf{U}$ in the M-step.

$$\max_\mathbf{U} \mathbb{E}_q\left[\log p(\mathbf{Y}, \mathbf{Z}, \mathbf{M}, \beta|\mathbf{U})p(\mathbf{U})\right] \tag{21}$$

Eliminating constant terms in the above equation, we need to solve the following optimization problem:

$$\begin{aligned}\max_\mathbf{U} f(\mathbf{U}) = & -n \log\det(\mathbf{K}) - \frac{1}{2}\mathrm{tr}(\mathbf{K}^{-1}\langle\mathbf{M}\rangle\mathbf{K}^{-1}\langle\mathbf{M}\rangle^\top) \\ & - \frac{1}{2}\mathrm{tr}(\mathbf{K}^{-1} \otimes \mathbf{K}^{-1} \cdot \mathbf{\Sigma}_\mathbf{M}) - \lambda\|\mathbf{U}\|_1\end{aligned} \tag{22}$$

Note that $\mathbf{K}$ is a shorthand for $\mathbf{K}(\mathbf{U}, \mathbf{U})$. The last term in (22) is a $l_1$ penalization term. Using standard matrix algebra technique, we can find the gradient of the first three terms in $f$ (i.e., omitting the $l_1$ penalization term) *w.r.t.* to a scalar $u_{i,r}$, the $r$-th element of $\mathbf{u}_i$,

$$\begin{aligned}\frac{\partial f}{\partial u_{ir}} = & -n\,\mathrm{tr}(\mathbf{K}^{-1}\frac{\partial \mathbf{K}}{\partial u_{ir}}) \\ & + \frac{1}{2}\mathrm{tr}\left[\mathbf{K}^{-1}\frac{\partial \mathbf{K}}{\partial u_{ir}}\mathbf{K}^{-1}(\overline{\mathbf{M}}\mathbf{K}^{-1}\overline{\mathbf{M}}^\top + \overline{\mathbf{M}}^\top\mathbf{K}^{-1}\overline{\mathbf{M}})\right] \\ & + \frac{1}{2}\mathrm{tr}(\mathbf{K}^{-1}\frac{\partial \mathbf{K}}{\partial u_{ir}}\mathbf{K}^{-1} \otimes \mathbf{K}^{-1} \cdot \mathbf{\Sigma}_\mathbf{M}) \\ & + \frac{1}{2}\mathrm{tr}(\mathbf{K}^{-1} \otimes \mathbf{K}^{-1}\frac{\partial \mathbf{K}}{\partial u_{ir}}\mathbf{K}^{-1} \cdot \mathbf{\Sigma}_\mathbf{M})\end{aligned} \tag{23}$$

To incorporate the $l_1$ penalty, we use a variant of the L-BFGS method to optimize $f(\mathbf{U})$ (Schmidt, 2010).

### 4.2 Efficient implementation

A naïve implementation of the above algorithm has a prohibitive $O(n^6)$ time complexity and a $O(n^4)$ space complexity at each iteration because of the expensive Kronecker product and the associated inversion. Now we present an efficient way to reduce the time complexity to $O(n^3)$ and the space complexity to $O(n^2)$.

Specifically we make use of the Kronecker product structures in the $n^2$ by $n^2$ covariance matrix $\mathbf{\Sigma_M}$. Let $\mathbf{K} = \mathbf{V}\Lambda\mathbf{V}^\top$ be the spectral decomposition of the old kernel matrix used to compute $\mathbf{\Sigma_M}$, where $\mathbf{V}$ is an orthogonal matrix and $\Lambda$ is nonnegative diagonal matrix. We have

$$\mathbf{\Sigma_M} = (\mathbf{V} \otimes \mathbf{V})\widetilde{\Lambda}(\mathbf{V} \otimes \mathbf{V})^\top \tag{24}$$

$$\widetilde{\Lambda} = \Lambda \otimes \Lambda(\mathbf{I} + \Lambda \otimes \Lambda)^{-1} \tag{25}$$

Note that $\widetilde{\Lambda}$ is also a diagonal matrix, and it can be computed from the diagonal elements of $\Lambda$. We define a square matrix $\mathbf{D}$ such that its entries are the diagonal elements of $\widetilde{\Lambda}$. If we denote the diagonal elements of $\Lambda$ by $\lambda_1, \ldots, \lambda_n$, then we have

$$D_{ij} = \frac{\lambda_i \lambda_j}{1 + \lambda_i \lambda_j} \tag{26}$$

According to Gupta and Nagar (2000), we can rewrite (14) as

$$\overline{\mathbf{M}} = \mathbf{V}[(\mathbf{V}^\top(\overline{\mathbf{Z}} - \overline{\mathbf{P}})\mathbf{V}) \circ \mathbf{D}]\mathbf{V}^\top \tag{27}$$

where $\circ$ denotes the Hadamard product, i.e, the entry-wise multiplication of matrices.

Direct computation of $\mathrm{tr}(\mathbf{K}^{-1} \otimes \mathbf{K}^{-1} \cdot \mathbf{\Sigma_M})$ used in the log-probability of the joint model and its gradient takes $O(n^4)$ time. In order to reduce the time cost, we use the properties of trace operator. Note that $\widetilde{\Lambda}$ only has non-zero elements on its diagonal entries, we can write

$$\begin{aligned} & \mathrm{tr}(\mathbf{K}^{-1} \otimes \mathbf{K}^{-1} \cdot \mathbf{\Sigma_M}) \\ =& \mathrm{tr}(\mathbf{K}^{-1} \otimes \mathbf{K}^{-1} \cdot (\mathbf{V} \otimes \mathbf{V})\widetilde{\Lambda}(\mathbf{V} \otimes \mathbf{V})^\top) \\ =& \mathrm{tr}((\mathbf{V}^\top\mathbf{K}^{-1}\mathbf{V}) \otimes (\mathbf{V}^\top\mathbf{K}^{-1}\mathbf{V}) \cdot \widetilde{\Lambda}) \\ =& \mathrm{diag}(\mathbf{V}^\top\mathbf{K}^{-1}\mathbf{V})^\top \cdot \mathbf{D} \cdot \mathrm{diag}(\mathbf{V}^\top\mathbf{K}^{-1}\mathbf{V}) \end{aligned} \tag{28}$$

where $\mathrm{diag}(\mathbf{A})$ is the column vector consists of the diagonal elements of a square matrix $\mathbf{A}$. We can similarly derive the gradient

$$\begin{aligned} & \frac{\partial}{\partial u_{ir}} \mathrm{tr}(\mathbf{K}^{-1} \otimes \mathbf{K}^{-1} \cdot \mathbf{\Sigma_M}) \\ =& 2\ \mathrm{diag}(\mathbf{V}^\top\mathbf{K}^{-1}\frac{\partial \mathbf{K}}{\partial u_{ir}}\mathbf{K}^{-1}\mathbf{V})^\top \cdot \mathbf{D} \cdot \mathrm{diag}(\mathbf{V}^\top\mathbf{K}^{-1}\mathbf{V}) \end{aligned} \tag{29}$$

Now we reduce the time complexity to $O(n^3)$ and the space complexity to $O(n^2)$ for each optimization iteration. We may further approximate the old covariance matrix $\mathbf{K}$ by the first $m$ leading eigenvalues and the corresponding eigenvectors. This approximation will further reduce the time complexity to $O(mn^2)$ if a Lanczos-type method is employed to find the kernel approximation.

# 5 Experiment

In this section, we demonstrate how SMGB performs on both synthetic data and real world networks compared with alternative methods. In all of our experiments, we use the isotropic kernel function for our SMGB model, *i.e.* $k(\mathbf{u}_i, \mathbf{u}_j) = e^{-\gamma\|\mathbf{u}_i - \mathbf{u}_j\|^2}$ The hyperparameter $\gamma$ is determined by cross-validation.

We compare the proposed SMGB model with the following models in our experiments.

- Mixed membership stochastic blockmodels (MMSBs) (Airoldi et al., 2008). MMSB focuses on modeling the membership assignment for each entity by a generative probabilistic model.

- Latent eigenmodels (LEM) (Hoff, 2007). LEM generalizes the distance model (Hoff et al., 2002) and the latent class model (Snijders and Nowicki, 1997). In practice, LEM has good interaction/link prediction ability.

For both models, we use the R code retrieved from The Comprehensive R Archive Network[1] (CRAN). Instead of variational-EM inference described in the original MMSB paper, the CRAN R code uses collapsed Gibbs sampling to learn the model.

## 5.1 Experiment on Synthetic Data

We generate the synthetic data in the following way. We first specify a $30 \times 30$ interaction matrix, representing a network with three 10-node cliques. In each clique the nodes are fully connected (so the corresponding sub-matrix is dense). We then randomly add or remove edges for 5% of all 900 possible interactions $(i, j)$ between all the entities. We use this noisy matrix as our observation $\mathbf{Y}$. For the synthetic data, we set the number of groups to $d = 3$ for all methods.

To examine how well does our method perform on interaction/link prediction tasks, we generated 10 random splits of training and test data of interations/non-interactions from the synthetic data with 80% of the matrix values as training data, and compute the averaged Area Under Curve (AUC) values and their standard error from $\mathbf{X}$ against the ground truth test data. The result is shown in Figure 1. We can see

[1] cran.r-project.org/

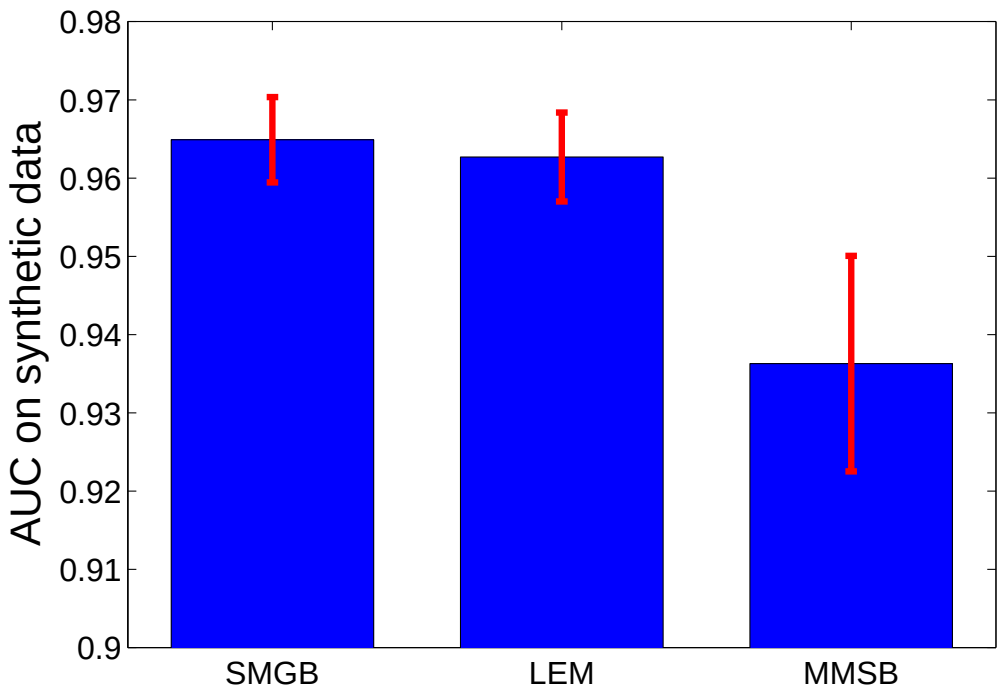

Figure 1: The AUC on the synthetic data. SMGB outperforms LEM and MMSB.

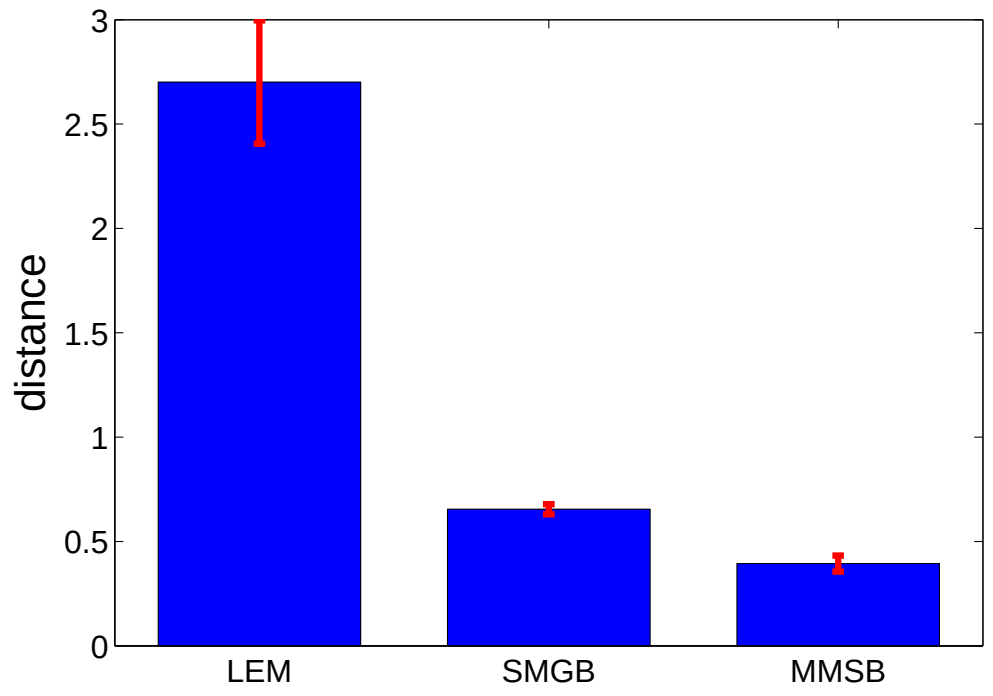

Figure 2: The average distance between the latent membership matrix $\mathbf{U}$ and the ground truth membership matrix $\mathbf{U}^0$. Standard error bar is plotted. SMGB is slightly worse, but comparable, than MMSB. Both methods are much better than LEM on membership discovery.

that our model significantly outperforms MMSB and is marginally better than LEM.

Next, we examine the latent membership vectors learned by each model. Let $\mathbf{e}_i$ be a vector whose $i^{th}$ entry is 1 and all other entries are 0. For MMSB, we assume the ground truth membership for entity $i$ is $\mathbf{u}_i^0 = \mathbf{e}_{c_i}$, if $i$ belongs to the group $c_i$. The goodness of $\mathbf{U}$ is measure by the distance $\|\mathbf{U} - \mathbf{U}^0\|_F$, where $\mathbf{U}^0$ is the ground truth membership matrix. The lower the distance, the more accurate the membership assignment. To compare across the models with this ground truth membership assignment, we constrain the optimizer to optimize over nonnegative values in SMGB , then we normalize each $\mathbf{u}_i$ such that the summation of all entries of $\mathbf{u}_i$ is equal to 1. There is no good way of obtaining interpretable membership assignments from the latent vector learned in LEM. We use the following heuristic: all negative values in $\mathbf{U}$ learned by LEM is set to 0, then we normalize each $\mathbf{u}_i$ to obtain the membership assignments and compute the distance to $\mathbf{U}^0$;

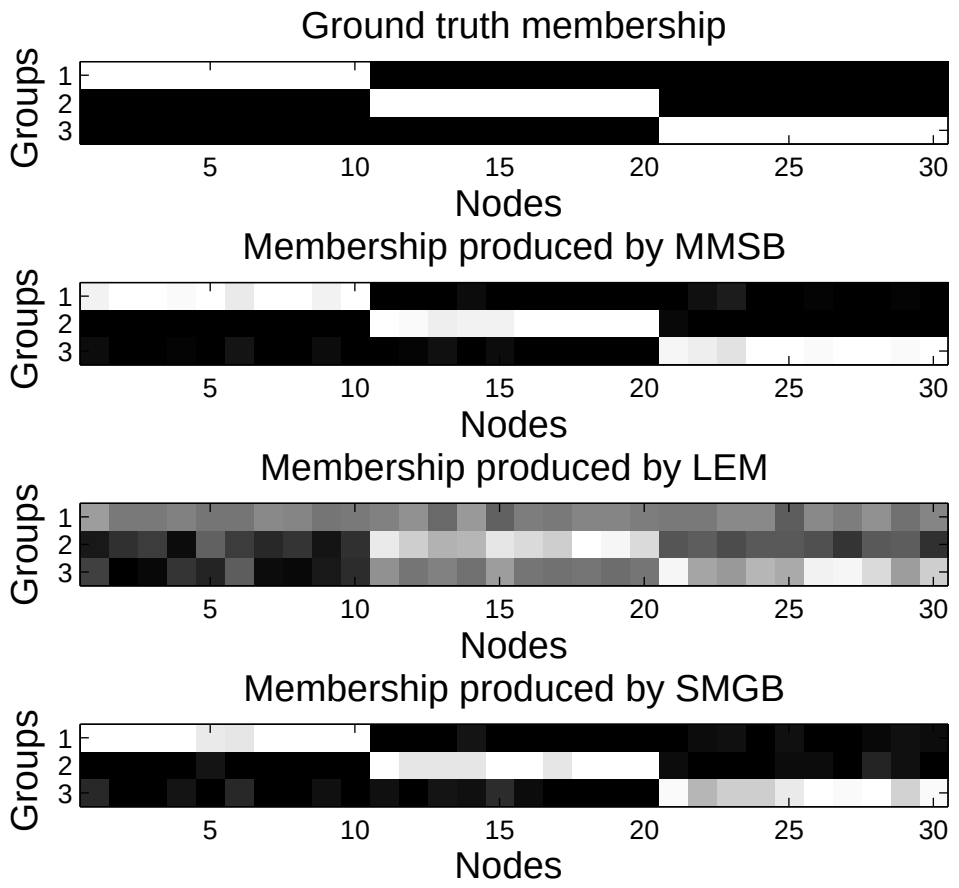

Figure 3: The structure of the ground truth $\mathbf{U}^0$ and latent membership matrix $\mathbf{U}$. SMGB achieves comparable quality as MMSB, and both are much better than LEM.

to compensate the negative value, we also do the same thing for $-\mathbf{U}$, and if its distance to $\mathbf{U}^0$ is smaller, we use the smaller distance.

We compute the distances for 10 runs with random noises. The averaged distance and standard error is shown in Figure 2. It is not surprised that MMSB has the smallest distance, because the model is designed to discover latent memberships. Our model, although worse than MMSB, is much better than that of LEM. These quantitative results are further confirmed by Figure 3, which shows the structures of ground truth $\mathbf{U}^0$ and $\mathbf{U}$ generated by the other three methods.

### 5.2 Experiment on Real-world Datasets

We employ two real-world datasets to test SMGB . It should be noted that the number of edges in a network is in the quadratic order of the number of nodes, and the prediction is made on each edge. The large number of edges makes the prediction task computationally challenging.

The network datasets are summarized in the following:

- The first dataset is a coauthorship data from the NIPS dataset, which are used in (Globerson et al., 2007). This dataset contains a list of all papers and authors from NIPS 1-17. We took the 100 authors who have published with the largest numbers of co-authors and keep the related edges.

- The second dataset represents friendship ties among 90 $12^{th}$-graders from the National Longitudinal Study of Adolescent Health [2]. The data is represented by a symmetric matrix corresponding to an undirected graph. $Y_{ij} = 1$ means nodes $i$ and $j$ are friends. This dataset is named as "Friends".

[2] `www.cpc.unc.edu/projects/addhealth`

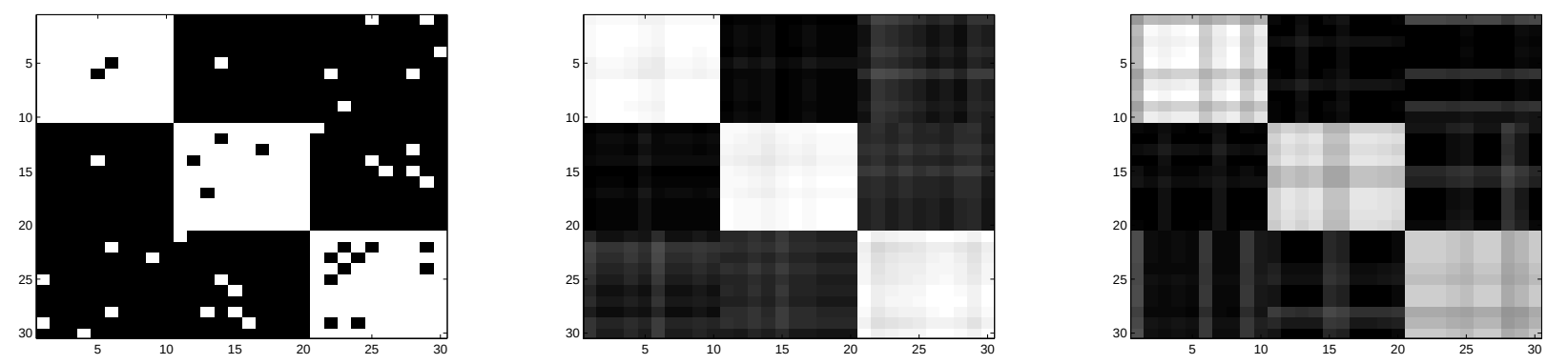

(a) (b) (c) (d)

Figure 4: The synthetic data, which is consists of 3 cliques. (a) Ground truth data with noises. (b)–(d) The posterior mean of $\mathbf{Y}$ by SMGB , MMSB and LEM respectively. SMGB and LEM reveal clearer structures than MMSB does.

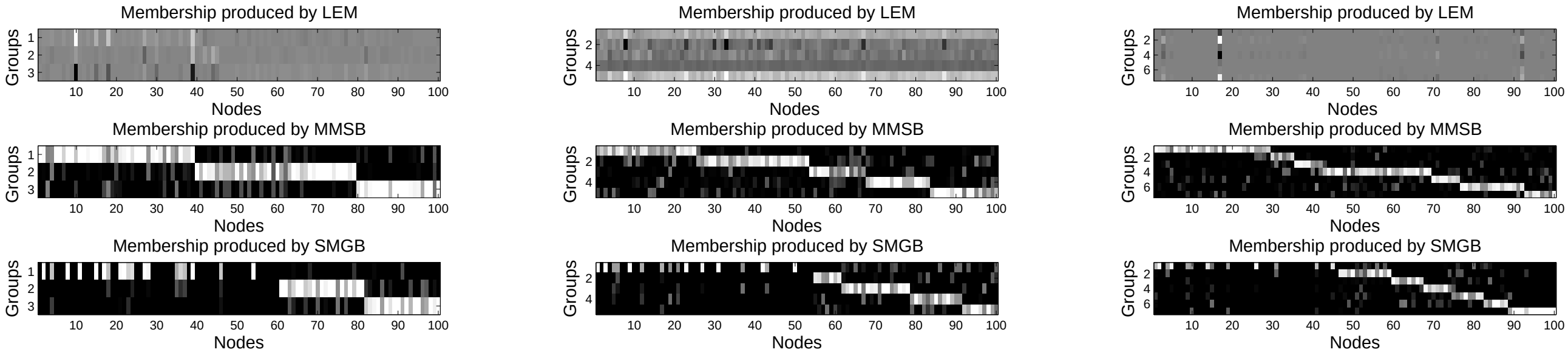

(a) *d=3, Coauthor* (b) *d=5, Coauthor* (c) *d=7, Coauthor*

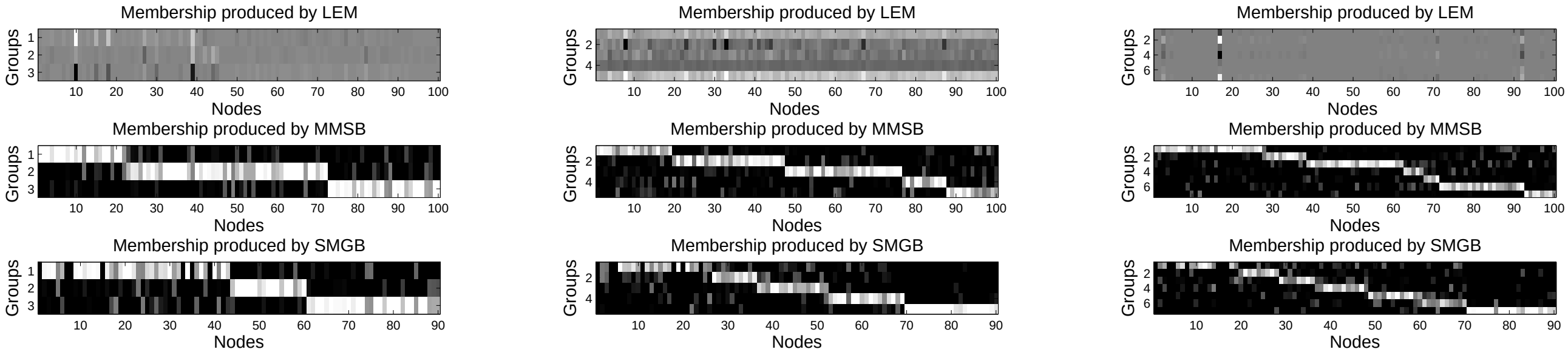

(d) *d=3, Friend* (e) *d=5, Friend* (f) *d=7, Friend*

Figure 5: The data plot and membership matrices of SMGB , MMSB and LEM on two relational datasets, Coauthor and Friend. The dimensions of the latent membership vector are $d = 3, 5, 7$ respectively. SMGB and MMSB show clear block structures, while LEM fails.

As the true group information is unknown in these network datasets, we vary the number of latent groups (i.e., the length of $\mathbf{u}_i$) from 3 to 5, and to 7. Since on these real network datasets, we do not have the ground truth information about latent memberships, we cannot not examine the membership estimation accuracy. So we compare all the models only on the prediction accuracy for hold out edges. Specifically for each of these datasets, we randomly choose 80% of the matrix elements (edges) for training and use the remaining for testing. The experiment is repeated 10 times. We evaluate all the models by Area Under Curve (AUC) values averaged over 10 runs. The higher an AUC value, the better the predictive performance. We report the average AUC values and the standard errors in Figure 6. SMGB outperforms all the other models in terms of the average AUC for all cases.

Figure 5 (a)-(c) and (d)-(f) show the visualization of the extracted membership structures for *Coauthor* and *Friend*, respectively. For each dataset, we vary the number of latent groups from 3 to 5, and to 7. Both SMGB and MMSB obtain clear group structures, while LEM fails.

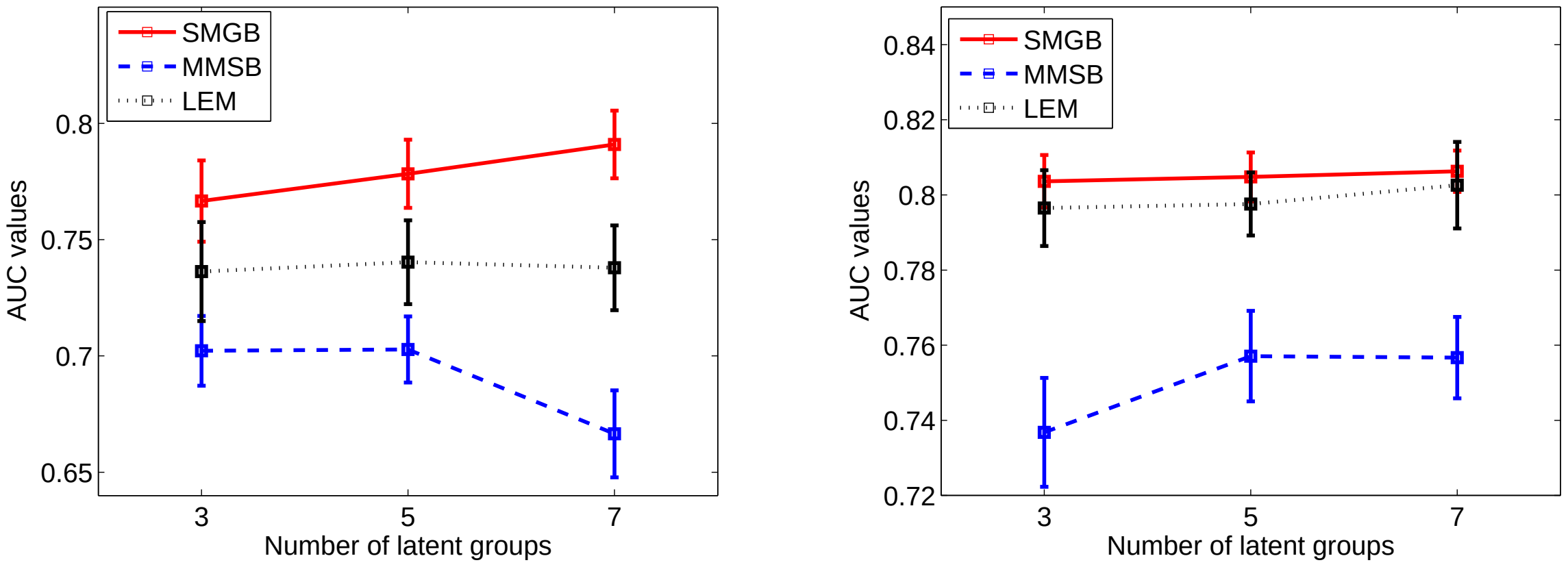

(a) *Coauthor* (b) *Friend*

Figure 6: The AUC values of SMGB , MMSB and LEM on Coauthor and Friend datasets. The dimensions of the latent membership vector are $d = 3, 5, 7$ respectively

## 6 Conclusions

In this paper, we have presented new nonparametric Bayesian relational models, Sparse Matrix-variate Gaussian process blockmodels, for discovering latent classes and modeling interactions of network nodes. The experimental results on synthetic and real network datasets demonstrate the advantages of SMGP over several state-of-the-art network models.

### Acknowledgement

This work was supported by NSF IIS-0916443 and NSF CAREER award IIS-1054903.

## References

Neil D. Lawrence and Raquel Urtasun. Non-linear matrix factorization with Gaussian processes. In *ICML '09: Proceedings of the 26th Annual International Conference on Machine Learning*, pages 601–608, 2009.

Edoardo M. Airoldi, David M. Blei, Stephen E. Fienberg, and Eric P. Xing. Mixed membership stochastic blockmodels. *J. Mach. Learn. Res.*, 9:1981–2014, June 2008.

Peter Hoff. Modeling homophily and stochastic equivalence in symmetric relational data. In *Advances in Neural Information Processing Systems 20*, 2007.

Tom A.B. Snijders and Krzysztof Nowicki. Estimation and prediction for stochastic blockmodels for graphs with latent block structure. *Journal of Classification*, 14(1): 75–100, 1997.

Charles Kemp, Thomas L. Griffiths, and Joshua B. Tenenbaum. Discovering latent classes in relational data. Technical Report AI Memo 2004-019, MIT, 2004.

Peter D. Hoff, Adrian E. Raftery, Mark S. Handcock, and Mark S. H. Latent space approaches to social network analysis. *Journal of the American Statistical Association*, 97:1090–1098, 2002.

Neil Lawrence. The Gaussian process latent variable model. Technical Report CS-06-03, The University of Sheffield, 2006.

Kai Yu and Wei Chu. Gaussian process models for link analysis and transfer learning. In *Advances in Neural Information Processing Systems 20*. 2008.

Zenglin Xu, Feng Yan, and Yuan Qi. Sparse matrix-variate $t$ process blockmodels. In *Proceeding of The 25th Artificial Intelligence (AAAI-11)*. 2011.

James H Albert and Siddhartha Chib. Bayesian analysis of binary and polychotomous response data. 88:669–679, 1993.

Arjun K. Gupta and D. K. Nagar. *Matrix variate distributions*. CRC Press, 2000.

Kai Yu, Wei Chu, Shipeng Yu, Volker Tresp, and Zhao Xu. Stochastic relational models for discriminative link prediction. In *Advances in Neural Information Processing Systems*, pages 333–340. MIT Press, 2007.

Mark Girolami and Simon Rogers. Variational Bayesian multinomial probit regression with Gaussian process priors. *Neural Computation*, 18:2006, 2005.

Mark Schmidt. *Graphical Model Structure Learning with L1-Regularization*. PhD thesis, University of British Columbia, 2010.

Amir Globerson, Gal Chechik, Fernando Pereira, and Naftali Tishby. Euclidean embedding of co-occurrence data. *Journal of Machine Learning Research*, 8:2265–2295, 2007.